\documentclass[conference]{IEEEtran}
\IEEEoverridecommandlockouts

\usepackage{cite}
\usepackage{amsmath,amssymb,amsfonts}
\usepackage{algorithmic}
\usepackage{graphicx}
\usepackage{textcomp}
\usepackage{xcolor}
\def\BibTeX{{\rm B\kern-.05em{\sc i\kern-.025em b}\kern-.08em
    T\kern-.1667em\lower.7ex\hbox{E}\kern-.125emX}}
\begin{document}

\title{Pre-training, Fine-tuning and Re-ranking: A Three-Stage Framework for Legal \\Question Answering}

\author{\IEEEauthorblockN{Shiwen Ni$^{1,*}$, Hao Cheng$^{1,*}$\thanks{$^*$These authors contributed equally.}, Min Yang$^{1,2,\dagger}$\thanks{$^\dagger$Corresponding author.}}
\IEEEauthorblockA{\textit{$^1$Shenzhen Key Laboratory for High Performance Data Mining} \\
\textit{Shenzhen Institutes of Advanced Technology, Chinese Academy of Sciences}\\
\textit{$^2$Shenzhen University of Advanced Technology} \\
sw.ni@siat.ac.cn; chenghao98@mail.ustc.edu.cn; min.yang@siat.ac.cn}
}
\maketitle

\begin{abstract}
Legal question answering (QA) has attracted increasing attention from people seeking legal advice, which aims to retrieve the most applicable answers from a large-scale database of question-answer pairs. Previous methods mainly use a dual-encoder architecture to learn dense representations of both questions and answers. However, these methods could suffer from lacking domain knowledge and sufficient labeled training data. In this paper, we propose a three-stage (\underline{p}re-training, \underline{f}ine-tuning and \underline{r}e-ranking) framework for \underline{l}egal \underline{QA} (called PFR-LQA), which promotes the fine-grained text representation learning and boosts the performance of dense retrieval with the dual-encoder architecture. Concretely, we first conduct domain-specific pre-training on legal questions and answers through a self-supervised training objective, allowing the pre-trained model to be adapted to the legal domain. Then, we perform task-specific fine-tuning of the dual-encoder on legal question-answer pairs by using the supervised learning objective, leading to a high-quality dual-encoder for the specific downstream QA task. Finally, we employ a contextual re-ranking objective to further refine the output representations of questions produced by the document encoder, which uses contextual similarity to increase the discrepancy between the anchor and hard negative samples for better question re-ranking.
We conduct extensive experiments on a manually annotated legal QA dataset. Experimental results show that our PFR-LQA method achieves better performance than the strong competitors for legal question answering. 
\end{abstract}

\begin{IEEEkeywords}
Legal question answering, Pre-training, Fine-tuning, Re-ranking
\end{IEEEkeywords}

\section{Introduction}\label{sec1}
A question answering (QA) system aims to retrieve the most appropriate answers from a large-scale question-answer database \cite{zhu2021retrieving}. With the ever-increasing size of legal cases in China, legal QA systems have attracted increasing attention from people seeking legal advice. The answer ranking (or question ranking) task is an important subtask of question answering, which aims to retrieve and re-rank relevant answers (or questions) in the question-answer database given a user query \cite{rao2017experiments,shen2018knowledge,chen2020re,matsubara2020reranking,kunneman2019question}. Early question answering methods mainly focused on designing various features \cite{tran2015jaist,severyn2013automatic}, such as syntactic features and dependency trees. However, the success of these methods relied heavily on feature engineering, which is a labor-intensive process. In recent years, deep learning models have become mainstream in the QA task and achieved impressive performance \cite{shen2018knowledge,chen2020re,matsubara2020reranking}. The promising results of deep neural networks come with the high costs of a large number of labeled training data \cite{ram2021few,abdel2022deep}. However, it is expensive to annotate accurate and sufficient labeled training samples for legal QA since legal expertise is needed to understand the legal documents and evaluate the relevance between each question-answer pair. 

Subsequently, pre-trained language models (PLMs) such as BERT \cite{kenton2019bert} and RoBERTa \cite{liu2019roberta} have achieved significant success in text retrieval and re-ranking by benefiting from rich knowledge derived from a large volume of unlabeled data \cite{glass2019span,ram2021few,jia-etal-2022-question}. For instance, \cite{jia-etal-2022-question} injected the question answering knowledge into a pre-training objective for learning general purpose contextual representations. 
Nevertheless, the PLMs pre-trained on the general corpus could lack legal knowledge and result in poor performance in legal QA. 

In this paper, we propose a three-stage (\underline{p}re-training, \underline{f}ine-tuning and \underline{r}e-ranking) framework for \underline{l}egal \underline{QA} (called PFR-LQA), which promotes the text representation learning and boosts the performance of the QA task with the dual-encoder architecture in the legal domain. 
Specifically, we first pre-train an encoder-decoder architecture tailored for dense retrieval on legal questions and answers through a self-supervised training objective. This can facilitate the dense retrieval model to be adapted to the legal domain effectively. Second, we fine-tune the pre-trained text retrieval model on the question-answer pairs through a standard supervised training objective, which can help the retrieval model learn high-quality text encoders for the QA task in the legal domain. Third, we employ a contextual re-ranking objective to further refine the output representations of questions produced by the dual-encoder, which uses contextual similarity to increase the discrepancy between the anchor and hard negative samples for learning better question representations. In this way, we can exploit the contextual similarity among the top-ranked questions with respect to a given query for better \textit{question similarity} that is the key task in question answering. After obtaining the semantically similar and relevant questions for a user query, we can return the corresponding answers to the retrieved questions as the recommended answers.

Our main contributions are three-fold. (1) We take advantage of the collaborative effect of domain-specific pre-training, task-specific fine-tuning and contextual re-ranking to learn better text representations for legal QA. (2) We invite legal practitioners (e.g., attorneys) to manually annotate a large-scale, high-quality LawQA dataset for legal QA. The release of them would push forward the research in this field. (3) Experimental results on the annotated LawQA dataset show that PFR-LQA achieves substantial improvements over the state-of-the-art baseline methods for legal QA.

\section{Methodology}

In this paper, we aim to retrieve and re-rank semantically similar and relevant  questions in the large-scale question-answer database for a given query. 
We propose a three-stage framework PFR-LQA for legal question answering. Specifically, we pre-train a dual-encoder model on legal questions and answers separately in the entire database $\mathcal{D}$ for better document representation learning ability. Second, we fine-tune the pre-trained dual-encoder model on question-answer pairs in $\mathcal{D}$. Third, we re-rank the candidate questions with query aggregation. 
Next, we will elaborate on these three modules.  

\begin{figure}[t]
	\centering
	\includegraphics[width=0.48\textwidth]{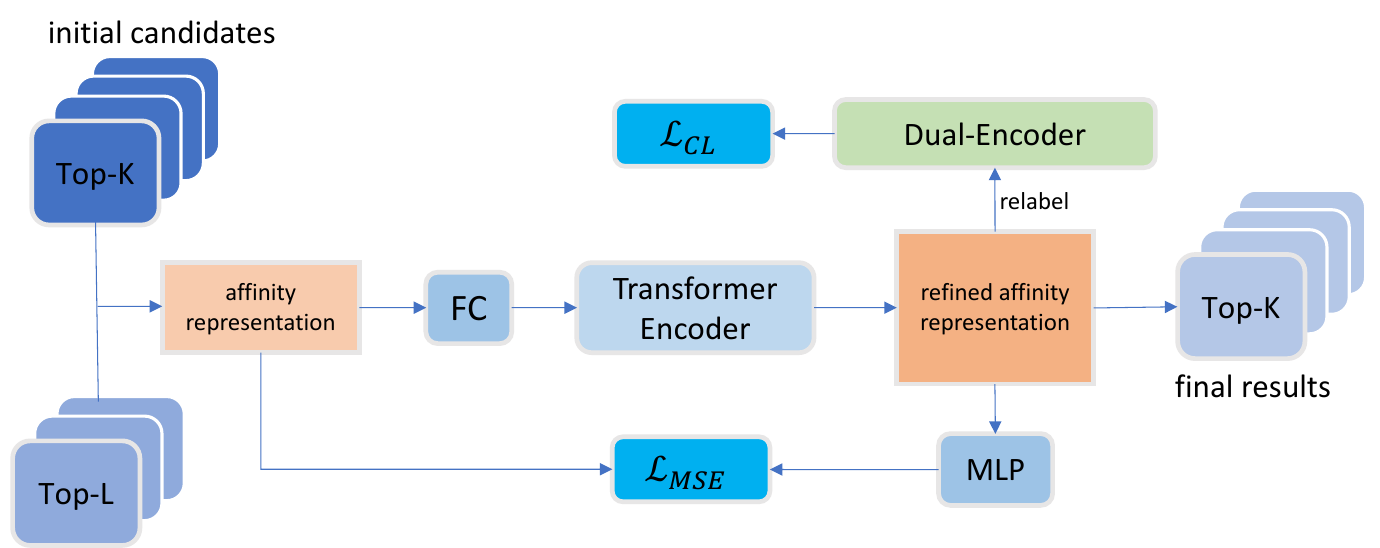}
	\caption{The training framework of contextual re-ranking.}
	\label{f1}
\end{figure}
\subsection{Pre-training on Legal Domain}
We pre-train an encoder-decoder architecture tailored for dense retrieval \cite{wu2022contextual} on the legal question and answers in the entire database $\mathcal{D}$. This domain-specific pre-training process helps to learn the semantics of each text span and the correlation between the text spans by leveraging self-supervised and context-supervised masked auto-encoding objectives, inspired by \cite{wu2022contextual}.
In particular, we first construct individual text spans via text segmentation and form each span pair by sampling two adjacent spans. On the encoder side, we apply a self-supervised pre-training to learn the semantics of each text span, which randomly masks out some tokens in a text span and then reconstructs the text span based on the unmasked tokens in the span. On the decoder side, we apply a context-supervised pre-training to learn the semantic correlations between a span pair, which reconstructs a text span in the pair based on its unmasked tokens and its adjacent span. Finally, the self-supervised and context-supervised pre-training objectives are combined to form a joint loss function $\mathcal{L}_{\rm pretrain}$ for pre-training. This pre-training model is optimized on the legal questions and answers separately in the entire database $\mathcal{D}$, which is denoted as Legal-SCP (\underline{Legal} \underline{S}elf-supervised and \underline{C}ontext-supervised \underline{P}re-training).   

\subsection{Fine-tuning on Legal QA}\label{sec:ft on LQA}
To improve the performance of the pre-trained model on the legal QA task, we further fine-tune the pre-trained model Legal-SCP on the question-answer pairs in the database $\mathcal{D}$. Following \cite{wu2022contextual}, we merely utilize the dual-encoder of Legal-SCP for question and answer encoding, while the decoder is discarded. That is, we use the dual-encoder of Legal-SCP as our base context encoder for learning the representations of questions and answers. 

Formally, we take the question $q_i = \{w_1^{q_i},..., w_n^{q_i}\}$ and the answer $a_i = \{w_1^{a_i},..., w_m^{a_i}\}$ separately as input of the pre-trained dual-encoder of Legal-SCP. Here, $n$ and $m$ represent the lengths of $q_i$ and $a_i$, respectively. We obtain the context representations of the question $q_i$ and the answer $a_i$ as follows:
\begin{equation}
	\begin{aligned}
		E^{q_i} &= Encoder\left(\left[{\rm CLS}, w_1^{q_i}, w_2^{q_i}, ..., w_n^{q_i}, {\rm SEP} \right]\right) \\
		E^{a_i} &= Encoder\left(\left[{\rm CLS}, w_1^{a_i}, w_2^{a_i}, ..., w_m^{a_i}, {\rm SEP} \right]\right) 
	\end{aligned}
\end{equation}
where the special token [CLS] is used to aggregate the sequence representation and the token [SEP] is used to separate the input text. For the sake of simplicity, We denote the hidden vector of the special [CLS] token as $E^{q_i}_{\rm CLS}$ (or $E^{a_i}_{\rm CLS}$) for the question $q_i$ (or the answer $a_i$), which can be considered as the base context representation of the question (or answer). 

Then, we calculate the correlation between the question $q_i$ and the answer $a_i$ by computing the cosine similarity between their corresponding context representations:
\begin{equation}
	{\rm sim}(q_i, a_i)={\rm cosine}(q_i, a_i) = \frac{E^{q_i}_{\rm CLS}\cdot E^{a_i}_{\rm CLS}}{||E^{q_i}_{\rm CLS}||\cdot ||E^{a_i}_{\rm CLS}||}
\end{equation}
where $\Vert.\Vert$ represents the L2 norm over contextual representations. 

To improve the correlation between the question and the answer, we fine-tune the dual-encoder using the legal questions and answers in $\mathcal{D}$. We adopt the circle loss \cite{sun2020circle} to optimize the dual-encoder model, which can maximize within-class similarity while minimizing between-class similarity, leading to improved deep feature representations. Specifically, we define the circle loss as follows:
\begin{equation}
	\small
	\mathcal{L}_2 = \log[1 + \sum_{(q_i, a_i^{+})\in \phi_{\rm pos}}\sum_{(q_i, a_i^{-})\in \phi_{\rm neg}}e^{\gamma({\rm sim}(q_i, a_i^{-}) - {\rm sim}(q_i, a_i^{+}) + m)}] \label{formula:3}
\end{equation}
where $\phi_{pos}$ and $\phi_{neg}$ represent the sets of positive and negative question-answer pairs with respect to the query $q_i$, respectively. $(q_i, a_i^{+})$ represents a positive question-answer pair and $(q_i, a_i^{-})$ represents a negative question-answer pair. $\gamma$ is a scale factor and $m$ is a margin factor that is used to improve the similarity separation. 

A crucial step of the above circle loss is to identify hard negative samples. In this work, we use the pre-trained Legal-SCP to retrieve a candidate set with top-$k$ answers for each query. These candidate answers excluding the labeled positive answers are treated as hard negative answers for the given query. 

\subsection{Question Re-ranking with Similarity Aggregation}
In practice, the queries are usually formulated using only few
terms, making it difficult to retrieve high-quality candidate answers.  
A potential solution to the above challenge is to explore the rich contextual similarity among the similar questions by refining each query with additional affinity features. 
Intuitively, if two queries are relevant, they should share similar distances from a set of anchor documents. 
To this end, we propose a query aggregation method with Transformer for contextual question re-ranking.

Given each query $q$, we first use the BM25 retrieval algorithm to extract top-$K$ similar questions, denoted as $R = \{q'_1, q'_2, ..., q'_K\}$. 
After fine-tuning the pre-trained Legal-SCP model on the legal QA corpus, we can use the pre-trained Legal-SCP model to learn a high-quality representation of each query $q$, denoted as $H^q$. We also use the fine-tuned Legal-SCP to learn the representation $H^{q'_j}$ for each candidate question $q'_j$.
We assume that the query $q$ is returned at the first position in the ranking list; if not, we insert it directly at the front of the ranking list. 
The overall representations of the candidate questions are denoted as $\mathcal{H}_K = \{H^{q'_1}, H^{q'_2},..., H^{q'_K}\} \in \mathbb{R}^{d_h \times K}$, where $d_h$ is the embedding dimension. 
To learn a better question representation, we select $L$ questions from the  candidate set $R$ as anchors for each query and calculate an affinity feature representation for each question to explore rich relational information contained in top-ranked questions. 
With anchor questions, we compute the affinity features for the $j$-th question (i.e., $q'_{j}$) in the rank list $R$ as follows:
\begin{equation}
	\small
	Q_j = [(H^{q'_j})^T H^{q'_1}, (H^{q'_j})^T H^{q'_2}, ..., (H^{q'_j})^T H^{q'_L}], 1 \le i \le L \le K
	\label{formula:4}
\end{equation}

Since the affinity representation $Q_j$ and the original question representation $H^{q'_j}$ are not in the same representation space, we first transform the affinity representation $Q_j \in \mathbb{R}^L$  into $Q_j' \in \mathbb{R}^{L'}$ with a learnable projection matrix $W_p$, which are then passed into a Transformer encoder as follows:
\begin{equation}
	Q_j' = Q_j W_p, \quad 
	Z_j = {\rm Transformer}(Q_j')
	\label{formula:5}
\end{equation}

We devise a contrastive loss $\mathcal{L}_{\rm CL}$ for better representation learning motivated by the intuition that relevant texts should have larger cosine similarity and vice versa. We define the contrastive loss $\mathcal{L}_{\rm CL}$ as follows:
\begin{equation}
	\small
	\mathcal{L}_{\rm CL} = \log[1 + \sum_{Z^{+}\in \delta_{\rm pos}}\sum_{Z^{-}\in \delta_{\rm neg}} e^{\gamma({\rm sim}(Z_1, Z^{-}) - {\rm sim}(Z_1, Z^{+}) + m)}]
\end{equation}
where $\delta_{\rm pos}$ (or $\delta_{\rm neg}$) denote the positive (or negative) samples that are related (not related) to the query in the candidate list $R$. Note that we use the fine-tuned dual-encoder retrieval model to decide the positive and negative samples by calculating question similarities. That is, the question pairs with small similarities (less than a threshold) are treated as negative samples, and the remaining retrieved candidate questions are regarded as positive samples.
$Z_1$ is the refined affinity representation of the query since we assume that the query is in the first position of the candidate list $R$. 

In order to retain the information contained in the original affinity representations, we apply a mean squared error (MSE) loss between the original affinity representation $Q_j$ and the refined affinity representation $Z_j$, which is defined as follows:
\begin{equation}
	\mathcal{L}_{\rm MSE} = \sum_{j=1}^K\Vert Q_j - \rho(Z_j) \Vert 
\end{equation}
where the $\rho(\cdot)$ function is a two-layer Multilayer Perceptron. 

Then, We minimize the joint loss function $\mathcal{L}_{\rm joint}$ by summing up the above two training objectives as:
\begin{equation}
	\mathcal{L}_{\rm joint} = \mathcal{L}_{\rm CL} + \lambda \mathcal{L}_{\rm MSE}
\end{equation}
Where $\lambda$ is a hyper-parameter that controls the relative importance of the MSE loss. Finally, we use the corresponding answers to
the re-ranked questions as the final recommended answers for each user query. 

\subsection{Inference stage}
During the inference phase, we first use the BM25 algorithm to retrieve top-$K$ questions for the input query. Second, we utilize the fine-tuned dual-encoder of Legal-SCP introduced in Section \ref{sec:ft on LQA} to produce the  representation $H^{q}$ for each query or candidate question $q$. Then, we learn a refined affinity representation $Z^{q}$ using Eq. (\ref{formula:4}) and Eq. (\ref{formula:5}) for  $q$. Finally, we calculate the correlation based on the refined representations of the user query and each question. The corresponding answers to
the re-ranked candidate questions are returned as the final recommended answers for the
given query. 

\section{Experiments }\label{sec4}
\section{Experimental Setup}
In this section, we will introduce experimental data, compared baselines and the implementation details. 

\subsection{Dataset}

We construct a large-scale Chinese QA dataset for evaluating the effectiveness of our PFR-LQA method. In particular, we collect real-life user queries from a Chinese law forum\footnote{http://china.findlaw.cn/}. We invite 15 professional lawyers to provide high quality answers to each user's query. We denote this Chinese QA dataset as LawQA, which is divided into nine categories, including \textit{Marriage and Family Affairs}, \textit{Labor Disputes}, \textit{Intellectual Property}, \textit{Criminal Offenses}, \textit{Property Disputes}, \textit{Corporate Compliance}, \textit{Urban Renewal}, \textit{Traffic Accidents}, and \textit{Medical Accidents}. With the help of professional lawyers, we collect a total of 549,668 positive QA pairs. Similar to InsuranceQA \cite{feng2015applying}, we randomly select several negative candidates via the BM25 algorithm from the training set for each positive QA pair. In total, there are 1,215,439 QA pairs for training, 97,414 QA pairs for validation, 900 QA pairs for testing.

\begin{table}[t]
	\small
	\centering
\caption{Statistics of the proposed LawQA dataset.}
		\setlength\tabcolsep{16pt} 
		\begin{tabular}{l | r }
			\hline
			Category & Number of Cases \\ \hline
			Marriage and Family Affairs & 59,189 \\ \hline
			Labor Disputes & 67,089 \\ \hline
			Intellectual Property & 16,622 \\ \hline
			Criminal Offenses & 90,966 \\ \hline
			Property Disputes & 21,720 \\ \hline
			Corporate Compliance & 22,497 \\ \hline
			Urban Renewal & 796 \\ \hline
			Traffic Accidents & 35,535 \\ \hline
			Medical Accidents & 8,061 \\ \hline
	\end{tabular}
\end{table}

\subsection{Baselines}
To verify the effectiveness of our proposed method, we compare our PFR-LQA method with several state-of-the-art baselines which are described below: BERT \cite{kenton2019bert}, LawFormer \cite{xiao2021lawformer}, DPR \cite{karpukhin-etal-2020-dense}, ColBERT \cite{khattab2020colbert}, SimCSE \cite{gao-etal-2021-simcse}.

\subsection{Implementation Details}
In the first stage (domain-specific pre-training), we initialize the encoder of Legal-SCP with RoBERTa-wwm-ext \cite{liu2019roberta}, while the decoder of Legal-SCP is trained from scratch. We adopt the QA pairs in the entire database as the pre-training data. 
We use HanLP \cite{he-choi-2021-stem} to split each answer into sentences and group consecutive sentences into spans, with a maximum span length of 128. The batch size is set to 54. The pre-training is optimized using the AdamW \cite{loshchilov2017decoupled} optimizer with an initial learning rate of 1e-4. A linear schedule is used with a warmup ratio of 0.1. 
In the second stage (task-specific fune-tuning), we discard the decoder of the pre-trained Legal-SCP, and only used the encoder for fine-tuning on the question answering task. The maximum sequence length is set to 128. The training batch size is set to 192. The dimension of the hidden state  (i.e., $h_d$) is set to 768. We adopt the AdamW optimizer for optimization, with an initial learning rate of 5e-5. For simplicity, the scale factor $\gamma$ and margin $m$ are set to 20 and 0, respectively. For each query, we retrieve the top-16 candidate answers for re-ranking by using the BM25 algorithm (i.e., $K=16$). 
In the third stage (contextual re-ranking), we set the number of anchor questions to 8 (i.e., $L=8$). The batch size is set to 256. We adopt the SGD optimizer \cite{bottou2012stochastic} for model optimization, with an initial learning rate of 0.1, a weight decay of $10^{-6}$, and a momentum of 0.9. We use a cosine scheduler to gradually decay the learning rate to 0. Similar to the second stage, the scale factor $\gamma$ and margin $m$ are set to 20 and 0, respectively. The hyperparameter $\lambda$ is set to 0.2. 
We measure our method for legal question answering using two official metrics on the test set: Precision at 1 of ranked candidates ($P$@1) and Mean Reciprocal Rank (MRR@16).

\subsection{Experimental Results}
Table \ref{tab:mainresults} summarizes the experimental results on our LawQA dataset. Our PFR-LQA outperforms the compared baselines significantly on the LawQA dataset, verifying the effectiveness of our PFR-LQA method. From Table \ref{tab:mainresults}, we can observe that the fine-tuned PLMs (e.g., BERT~\cite{kenton2019bert} and RoBERTa~\cite{liu2019roberta}) perform poorly because these methods do not incorporate the legal knowledge in the pre-training stage. LawFormer \cite{xiao2021lawformer}, which is pre-trained on long legal documents, outperforms BERT and RoBERTa. This indicates the domain-specific pre-training can improve the performance of legal question answering to a certain extent. DPR \cite{karpukhin-etal-2020-dense}, ColBERT \cite{khattab2020colbert} and SimCSE \cite{gao-etal-2021-simcse} further improve the performance of LawFormer through the usage of dual-encoder to learn high-quality question and answer representations. Our PFR-LQA method takes a further step towards learning better question and answer representations for question answering by taking the collaborative effect of domain-specific pre-training, task-specific fine-tuning and contextual re-ranking. In particular, PFR-LQA outperforms the best-performing baseline by 5.7\% on P@1 and 4.2\% on MRR@16.
\begin{table}[t]
	\centering
	\caption{Experimental results in terms of P@1 and MRR on LawQA.}
	\label{tab:mainresults}
	\scalebox{1}{
		\setlength\tabcolsep{20pt} 
		\begin{tabular}{l|c|c}
			\hline
			\textbf{Model} & \textbf{P@1} & \textbf{MRR@16} \\
			\hline
			BM25 & 59.3 & 72.5 \\
			BERT & 67.6 & 79.2 \\
			RoBERTa & 68.6 & 80.1 \\
			LawFormer & 69.9 & 81.4 \\
			DPR & 72.8 & 81.4 \\
			ColBERT & 71.0 & 80.9 \\
			SimCSE & 74.2 & 83.1 \\
			\textbf{PFR-LQA} (Ours) & \textbf{79.9} & \textbf{87.3} \\\hline
			\quad \textit{w/o} DSP & 75.8 & 84.5 \\
			\quad \textit{w/o} TSF & 73.7 & 82.8 \\
			\quad \textit{w/o} CR & 78.5 & 86.5 \\
			\hline
		\end{tabular}
	}
\end{table}

\subsection{Ablation Study}
To verify the effectiveness of each component in PFR-LQA, we peform ablation test of PFR-LQA on the test set of LawQA in terms of removing the domain-specific pre-training (denoted as w/o DSP), task-specific fine-tuning (denoted as w/o TSF), and contextual re-ranking (denoted as w/o CR), respectively. We summarize the ablation test results in Table \ref{tab:mainresults}. From the results, we can observe that the task-specific fine-tuning module has the largest impact on the performance of PFR-LQA. This is because it can help learn high-quality question and answer representations. The improvement of the domain-specific pre-training is also significant since the pre-training in the legal domain is the basis of the latter fine-tuning and re-ranking processes. It is no surprise that combining all the factors achieves the best performance on the LawQA.


\section{Conclusion}\label{sec6}
We present a three-stage framework for legal question-answering (referred to as PFR-LQA), which emphasizes the importance of fine-grained text representation learning and enhances the performance of dense retrieval through a dual-encoder architecture. In the first stage, we carry out domain-specific pre-training on general legal documents using a self-supervised training objective, which allows the pre-trained language model (PLM) to be customized for the legal domain. In the second stage, we perform task-specific fine-tuning of the document encoder on legal question-answering pairs using a standard supervised learning objective, resulting in high-quality encoders that are well-suited for the specific downstream task. Finally, we utilize a contextual re-ranking objective to further refine the output representations generated by the dual-encoders. This objective employs contextual similarity to increase the differentiation between the anchor and hard negative samples for better representation learning. To evaluate the effectiveness of PFR-LQA, we conduct extensive experiments on the LawQA. The results demonstrate that our proposed method outperforms competitive approaches in legal question answering, showcasing its superior performance.

\section{Acknowledgements}
This work was supported by National Natural Science Foundation of China (62376262), Natural Science Foundation of Guangdong Province of China (2024A1515030166), Shenzhen Science and Technology Innovation Program (KQTD20190929172835662), GuangDong Basic and Applied Basic Research Foundation (2023A1515110718 and 2024A1515012003), China Postdoctoral Science Foundation (2024M753398), Postdoctoral Fellowship Program of CPSF (GZC20232873).
\bibliographystyle{IEEEtran}
\bibliography{IEEEfull}

\begin{thebibliography}{10}
\providecommand{\url}[1]{#1}
\csname url@samestyle\endcsname
\providecommand{\newblock}{\relax}
\providecommand{\bibinfo}[2]{#2}
\providecommand{\BIBentrySTDinterwordspacing}{\spaceskip=0pt\relax}
\providecommand{\BIBentryALTinterwordstretchfactor}{4}
\providecommand{\BIBentryALTinterwordspacing}{\spaceskip=\fontdimen2\font plus
\BIBentryALTinterwordstretchfactor\fontdimen3\font minus
  \fontdimen4\font\relax}
\providecommand{\BIBforeignlanguage}[2]{{%
\expandafter\ifx\csname l@#1\endcsname\relax
\typeout{** WARNING: IEEEtran.bst: No hyphenation pattern has been}%
\typeout{** loaded for the language `#1'. Using the pattern for}%
\typeout{** the default language instead.}%
\else
\language=\csname l@#1\endcsname
\fi
#2}}
\providecommand{\BIBdecl}{\relax}
\BIBdecl

\bibitem{zhu2021retrieving}
F.~Zhu, W.~Lei, C.~Wang, J.~Zheng, S.~Poria, and T.-S. Chua, ``Retrieving and
  reading: A comprehensive survey on open-domain question answering,''
  \emph{arXiv preprint arXiv:2101.00774}, 2021.

\bibitem{rao2017experiments}
J.~Rao, H.~He, and J.~Lin, ``Experiments with convolutional neural network
  models for answer selection,'' in \emph{Proceedings of the 40th International
  ACM SIGIR Conference on Research and Development in Information Retrieval},
  2017, pp. 1217--1220.

\bibitem{shen2018knowledge}
Y.~Shen, Y.~Deng, M.~Yang, Y.~Li, N.~Du, W.~Fan, and K.~Lei, ``Knowledge-aware
  attentive neural network for ranking question answer pairs,'' in \emph{The
  41st International ACM SIGIR Conference on Research \& Development in
  Information Retrieval}, 2018, pp. 901--904.

\bibitem{chen2020re}
D.~Chen, S.~Peng, K.~Li, Y.~Xu, J.~Zhang, and X.~Xie, ``Re-ranking answer
  selection with similarity aggregation,'' in \emph{Proceedings of the 43rd
  International ACM SIGIR Conference on Research and Development in Information
  Retrieval}, 2020, pp. 1677--1680.

\bibitem{matsubara2020reranking}
Y.~Matsubara, T.~Vu, and A.~Moschitti, ``Reranking for efficient
  transformer-based answer selection,'' in \emph{Proceedings of the 43rd
  international ACM SIGIR conference on research and development in information
  retrieval}, 2020, pp. 1577--1580.

\bibitem{kunneman2019question}
F.~Kunneman, T.~C. Ferreira, E.~Krahmer, and A.~Van Den~Bosch, ``Question
  similarity in community question answering: A systematic exploration of
  preprocessing methods and models,'' in \emph{Proceedings of the International
  Conference on Recent Advances in Natural Language Processing (RANLP 2019)},
  2019, pp. 593--601.

\bibitem{tran2015jaist}
Q.~H. Tran, D.-V. Tran, T.~Vu, M.~Le~Nguyen, and S.~B. Pham, ``Jaist: Combining
  multiple features for answer selection in community question answering,'' in
  \emph{Proceedings of the 9th International Workshop on Semantic Evaluation
  (SemEval 2015)}, 2015, pp. 215--219.

\bibitem{severyn2013automatic}
A.~Severyn and A.~Moschitti, ``Automatic feature engineering for answer
  selection and extraction,'' in \emph{Proceedings of the 2013 Conference on
  Empirical Methods in Natural Language Processing}, 2013, pp. 458--467.

\bibitem{ram2021few}
O.~Ram, Y.~Kirstain, J.~Berant, A.~Globerson, and O.~Levy, ``Few-shot question
  answering by pretraining span selection,'' \emph{arXiv preprint
  arXiv:2101.00438}, 2021.

\bibitem{abdel2022deep}
H.~Abdel-Nabi, A.~Awajan, and M.~Z. Ali, ``Deep learning-based question
  answering: a survey,'' \emph{Knowledge and Information Systems}, pp. 1--87,
  2022.

\bibitem{kenton2019bert}
J.~D. M.-W.~C. Kenton and L.~K. Toutanova, ``Bert: Pre-training of deep
  bidirectional transformers for language understanding,'' in \emph{Proceedings
  of NAACL-HLT}, 2019, pp. 4171--4186.

\bibitem{liu2019roberta}
Y.~Liu, M.~Ott, N.~Goyal, J.~Du, M.~Joshi, D.~Chen, O.~Levy, M.~Lewis,
  L.~Zettlemoyer, and V.~Stoyanov, ``Roberta: A robustly optimized bert
  pretraining approach,'' \emph{ICLR}, 2020.

\bibitem{glass2019span}
M.~Glass, A.~Gliozzo, R.~Chakravarti, A.~Ferritto, L.~Pan, G.~Bhargav, D.~Garg,
  and A.~Sil, ``Span selection pre-training for question answering,''
  \emph{arXiv preprint arXiv:1909.04120}, 2019.

\bibitem{jia-etal-2022-question}
R.~Jia, M.~Lewis, and L.~Zettlemoyer, ``Question answering infused pre-training
  of general-purpose contextualized representations,'' in \emph{Findings of the
  Association for Computational Linguistics}, 2022, pp. 711--728.

\bibitem{wu2022contextual}
X.~Wu, G.~Ma, M.~Lin, Z.~Lin, Z.~Wang, and S.~Hu, ``Contextual mask
  auto-encoder for dense passage retrieval,'' \emph{AAAI}, 2023.

\bibitem{sun2020circle}
Y.~Sun, C.~Cheng, Y.~Zhang, C.~Zhang, L.~Zheng, Z.~Wang, and Y.~Wei, ``Circle
  loss: A unified perspective of pair similarity optimization,'' in
  \emph{Proceedings of the IEEE/CVF conference on computer vision and pattern
  recognition}, 2020, pp. 6398--6407.

\bibitem{feng2015applying}
M.~Feng, B.~Xiang, M.~R. Glass, L.~Wang, and B.~Zhou, ``Applying deep learning
  to answer selection: A study and an open task,'' in \emph{2015 IEEE workshop
  on automatic speech recognition and understanding (ASRU)}.\hskip 1em plus
  0.5em minus 0.4em\relax IEEE, 2015, pp. 813--820.

\bibitem{xiao2021lawformer}
C.~Xiao, X.~Hu, Z.~Liu, C.~Tu, and M.~Sun, ``Lawformer: A pre-trained language
  model for chinese legal long documents,'' \emph{AI Open}, vol.~2, pp. 79--84,
  2021.

\bibitem{karpukhin-etal-2020-dense}
V.~Karpukhin, B.~Oguz, S.~Min, P.~Lewis, L.~Wu, S.~Edunov, D.~Chen, and W.-t.
  Yih, ``Dense passage retrieval for open-domain question answering,'' in
  \emph{Proceedings of the 2020 Conference on Empirical Methods in Natural
  Language Processing}, 2020, pp. 6769--6781.

\bibitem{khattab2020colbert}
O.~Khattab and M.~Zaharia, ``Colbert: Efficient and effective passage search
  via contextualized late interaction over bert,'' in \emph{Proceedings of the
  43rd International ACM SIGIR conference on research and development in
  Information Retrieval}, 2020, pp. 39--48.

\bibitem{gao-etal-2021-simcse}
T.~Gao, X.~Yao, and D.~Chen, ``{S}im{CSE}: Simple contrastive learning of
  sentence embeddings,'' in \emph{Proceedings of the 2021 Conference on
  Empirical Methods in Natural Language Processing}, 2021, pp. 6894--6910.

\bibitem{he-choi-2021-stem}
H.~He and J.~D. Choi, ``The stem cell hypothesis: Dilemma behind multi-task
  learning with transformer encoders,'' in \emph{Proceedings of the 2021
  Conference on Empirical Methods in Natural Language Processing}, 2021, pp.
  5555--5577.

\bibitem{loshchilov2017decoupled}
I.~Loshchilov and F.~Hutter, ``Decoupled weight decay regularization,''
  \emph{arXiv preprint arXiv:1711.05101}, 2017.

\bibitem{bottou2012stochastic}
L.~Bottou, ``Stochastic gradient descent tricks,'' \emph{Neural Networks:
  Tricks of the Trade: Second Edition}, pp. 421--436, 2012.

\end{thebibliography}

\end{document}